\documentclass[letterpaper, 10 pt, conference]{ieeeconf}  

\IEEEoverridecommandlockouts                              

\input{header}

\usepackage{newtxtext} 
\usepackage{newtxmath} 
\usepackage{microtype}
\usepackage{times} 

\title{ \LARGE \bf \emph{FruitNeRF}: A Unified Neural Radiance Field based \\Fruit Counting Framework\\

\author{Lukas Meyer\textsuperscript{1}, Andreas Gilson\textsuperscript{2,3}, Ute Schmid\textsuperscript{3} and Marc Stamminger\textsuperscript{1}%
\thanks{The authors from $^{1}$ are with Visual Computing Erlangen (VCE), Friedrich-Alexander-Universität Erlangen-Nürnberg-Fürth, Germany, $^{2}$ is with the Fraunhofer Institute for Integrated Circuits (IIS) - EZRT, Fürth, Germany and \textsuperscript{3} are with Cognitive Systems, University of Bamberg, Germany   E-Mail:
 {\tt\small [lukas.meyer, marc.stamminger]@fau.de, andreas.gilson.fraunhofer.iis.de, ute.schmid@uni-bamberg.de}}%
}

}

\begin{document}

\maketitle



\begin{abstract}
We introduce \emph{FruitNeRF}, a unified novel fruit counting framework that leverages state-of-the-art view synthesis methods to count any fruit type directly in 3D. 
Our framework takes an unordered set of posed images captured by a monocular camera and segments fruit in each image. 
To make our system independent of the fruit type, we employ a foundation model that generates binary segmentation masks for any fruit. 
Utilizing both modalities, RGB and semantic, we train a semantic neural radiance field. 
Through uniform volume sampling of the implicit Fruit Field, we obtain fruit-only point clouds. 
By applying cascaded clustering on the extracted point cloud, our approach achieves precise fruit count.
The use of neural radiance fields provides significant advantages over conventional methods such as object tracking or optical flow, as the counting itself is lifted into 3D. 
Our method prevents double counting fruit and avoids counting irrelevant fruit.
We evaluate our methodology using both real-world and synthetic datasets. The real-world dataset consists of three apple trees with manually counted ground truths, a benchmark apple dataset with one row and ground truth fruit location, while the synthetic dataset comprises various fruit types including apple, plum, lemon, pear, peach, and mango.
Additionally, we assess the performance of fruit counting using the foundation model compared to a U-Net. 

\end{abstract}

\section{Introduction}
Due to a steadily growing global population~\cite{UN_ShiftingDemographics}, a declining workforce in several industrialized nations, and the advancement of climate change~\cite{PAEU}, the field of Precision Agriculture (PA) has seen a significant increase in both, application and research, in recent years~\cite{PAdevelopment}.

\begin{figure}[t!]
    \centering
    \includegraphics[width=1\linewidth]{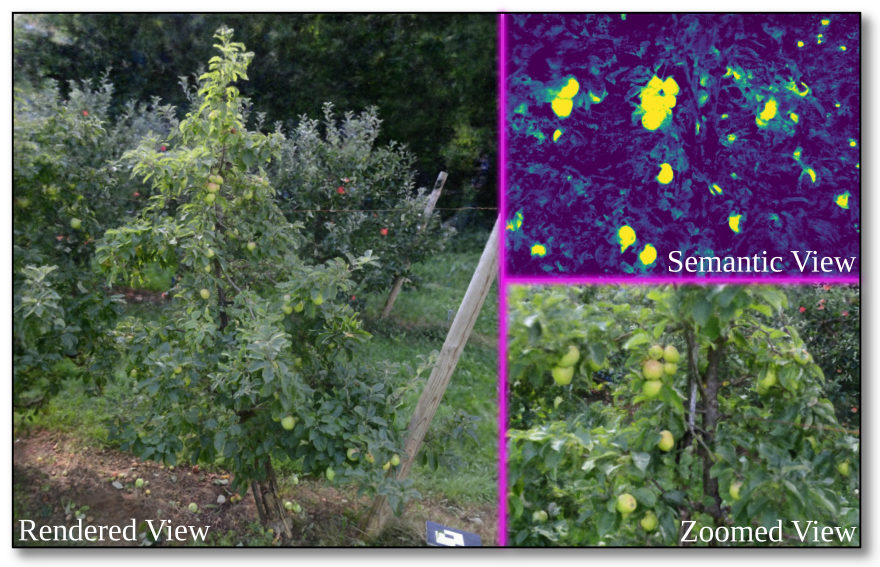}
\caption[]{Rendering of an apple tree from our real-world dataset generated with \emph{FruitNeRF}. On the right: a zoomed-in region with corresponding semantic logits (top) and the appearance rendering (bottom).}
\label{fig:Teaser}

\end{figure}

In PA, fruit counting is crucial for obtaining precise yield estimates to optimize harvest, and post-harvest management \cite{fcsota}. 
However, fruit counting remains a challenging task due to the need for accurate detection and tracking of fruits across multiple images \cite{liu2018robust} or in combination with 3D point clouds \cite{RoyCount}, regardless of visibility issues, partial occlusion, or varying lighting conditions.
In this context, it is crucial to prevent double counting and ensure that irrelevant fruit, such as fallen fruit or background fruit, are not erroneously included in the count.
Additionally, it is challenging to use the same counting method across different fruit types and environments. 


In our work, we propose \emph{FruitNeRF}, a novel unified fruit counting framework based on Neural Radiance Fields (NeRF) \cite{nerf}. The chosen architecture is inherently agnostic to the type of fruit and thus, provides the technical foundation for a generalized fruit counting approach. 

In the first phase, semantic image masks are calculated specific to the type of fruit under consideration. Combining the foundation models DINO \cite{dino} and Segment Anything (SAM) \cite{SAM}, fruit masks for all posed images are generated.
In comparison, we examine a specialized neural network, U-Net \cite{unet}, that was trained specifically on apples.
The subsequent step of our framework involves optimizing a semantic neural radiance field denoted as \emph{FruitNeRF}, utilizing both RGB and semantic masks to encode the spatial information of fruits within a neural radiance field.
In the third stage, we uniformly sample the density and semantic fields (rendered in Fig. \ref{fig:Teaser}) of the NeRF to acquire a point cloud that exclusively captures 3D points attributed to fruits.
In the last stage, the fruit point cloud is clustered, resulting in a precise fruit count.

\begin{figure*}[ht]
    \centering
    \includegraphics[width=1\linewidth]{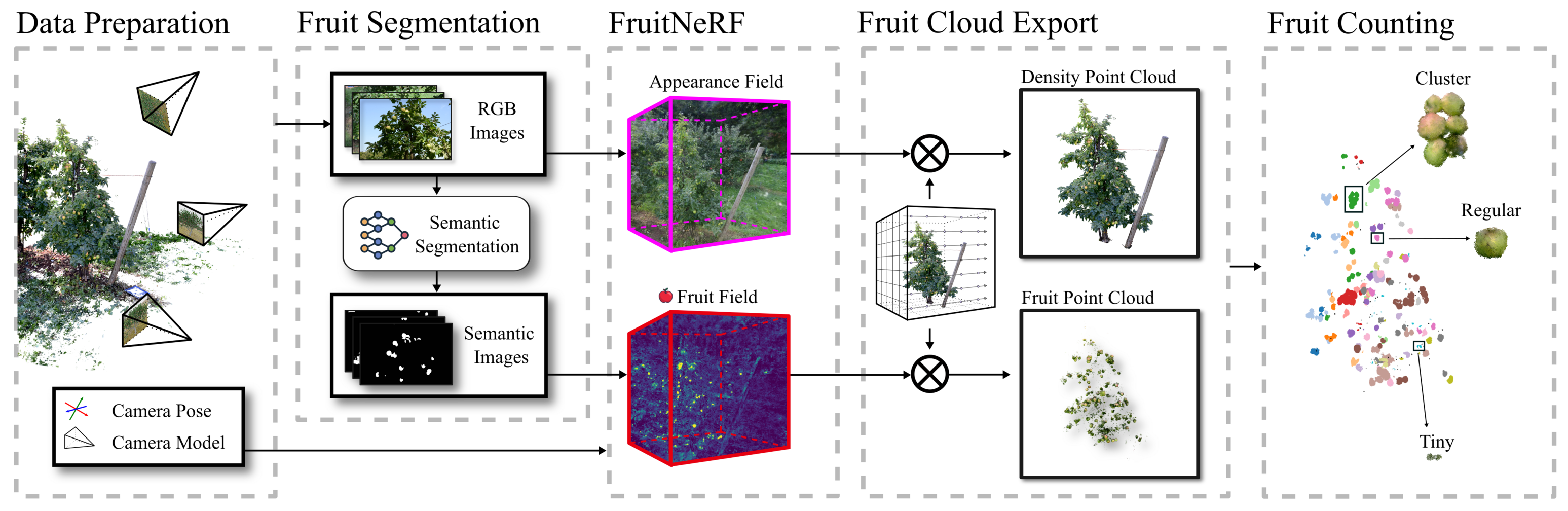}
\caption[]{Pipeline of our proposed fruit counting method - \emph{FruitNeRF}. Data Preparation (Sec.~\ref{sec:datapreparation}) uses structure from motion (SfM) \cite{schoenberger2016sfm} to recover both intrinsic and extrinsic camera parameters. We then extract semantic masks for arbitrary fruit types (Sec.~\ref{ssec:fruit_segmentation}) using the foundation model SAM \cite{SAM} and a self-trained U-Net \cite{unet} for apples only.
The posed RGB and semantic images are used to train a semantic neural radiance field. \emph{FruitNeRF} (Sec.~\ref{sec:fruit_nerf}) encodes the appearance of the scene including the semantic information. By sampling the appearance and Fruit Field (Sec.~\ref{sec:nerf_sampling}) uniformly, a dense point cloud is obtained. This paves the way for selecting only the 3D fruit points and clustering them to achieve a precise fruit count (Sec.~\ref{sec:fruit_count}).}
\label{fig:FruitNeRF_Pipeline}
\end{figure*}

We evaluate our framework with both synthetic and real-world data and demonstrate that \emph{FruitNeRF} generalizes well across different fruit types.
The main contributions of our work are:

\begin{itemize}
    \item We propose a novel fruit counting method from unordered images utilizing semantic NeRFs.
    \item We release a fruit dataset comprising synthetic data from various fruit trees and a real-world dataset specifically focused on apple trees\footnote{Project website: \url{https://meyerls.github.io/fruit_nerf}}.
    \item The code of \emph{FruitNeRF}\footnote{FruitNeRF code: \url{https://github.com/meyerls/FruitNeRF}} has been made open-source.
\end{itemize}

\section{Related Work}
Recent advances in computer vision and hardware have made monitoring fruits a feasible task in horticulture. A strong focus in this connection is laid on sweet peppers \cite{pagnerf, pathobot}, strawberries \cite{hqstrawberry, berryharvest}, tomatoes \cite{pathobot}, and apples \cite{liu2018robust, liu2019monocular, hanI2020comparative, FroutCountingCNN}. Therefore, we present a short overview of fruit counting methods.
In literature, fruit counting pipelines are commonly split into two distinct tasks: fruit detection, and fruit tracking and counting.

While the detection stage was dominated by hand-crafted features (e.g., color-based, shaped-based, etc.) \cite{fcsota} in the past, with the rise of deep learning this task has been successfully replaced by a vast variety of segmentation or detection network architectures (e.g., VCC, ResNet, YOLO) \cite{fcsota}. 

Fruit tracking and counting on the other hand are still challenging. 
The main causes of errors are double counting and counting fruit outside the region of interest (e.g., fallen fruit or fruit from trees in back rows).
Especially, double counting arises from various sources, such as observing the same fruit in two consecutive images or counting fruit from both sides.
To address this issue, researchers have proposed the following strategies \cite{fcsota}: counting fruits only on a per image level \cite{DataDrivenCounting}, tracking fruits across successive frames \cite{liu2018robust, liu2019monocular}, leveraging sparse point cloud data to count the fruit location in space \cite{FroutCountingCNN, RoyCount, hanI2020comparative} or projecting 2D instance segmentation onto 3D space \cite{fuji_count}.

The work from Liu \emph{et al.} \cite{liu2018robust} is the first to cover the entire automatic fruit counting pipeline applied to an image sequence. 
In the detection phase, they employ a network architecture, which segments each image into fruit and non-fruit pixels. 
To assign masks across multiple consecutive frames, they use a Kalman Filter-corrected optical flow tracker. 
Additionally, they localize fruit locations in 3D by tracking image features across frames to rectify counting errors such as double counting or detecting background and ground fruit.

Häni \emph{et al.} \cite{hanI2020comparative} present a modular end-to-end counting system in apple orchards, which connects multiple components from previous work regarding fruit detection \cite{FroutCountingCNN}, counting \cite{FroutCountingCNN} and tracking \cite{RoyCount}. 
In their approach, fruit detection and counting are merged by determining the fruit number on a per-image level. 
They compute a semantic point cloud by projecting the point cloud back to all camera frames and computing the intersection with the segmentation mask to identify 3D points belonging to apples \cite{RoyCount}. 
The point clouds are afterward clustered and projected clusterwise to the image plane to obtain the number of apples in a cluster \cite{FroutCountingCNN}.

Gené-Mola \emph{et al.} \cite{fuji_count} compute the instance segmentation mask for all images using a convolutional neural network. To obtain a semantic point cloud they use Structure from Motion (SfM) \cite{schoenberger2016sfm} with masked images and to cluster the corresponding point cloud. By back-projecting the clusters into multiple images they assign each cluster an instance ID.
 
However, the discussed approaches \cite{liu2018robust}, \cite{hanI2020comparative}, and \cite{fuji_count} are all combining SfM and 2D segmentation masks to compute the semantic point cloud and the position of the apples in space. 
In comparison with \emph{FruitNeRF}, we first perform a semantic reconstruction and then perform the fruit counting in 3D resulting in improved reliability. 

\section{Proposed Approach}

In this chapter, we introduce the methodology and pipeline for \emph{FruitNeRF} which is depicted in Fig. \ref{fig:FruitNeRF_Pipeline}.
\subsection{Data Preparation}
\label{sec:datapreparation}
The initial step for our pipeline is data preparation. Both our synthetic and real-world datasets consist of sets of RGB images.
A detailed description of the generated and recorded data is provided in section \ref{sec:dataset}. 

For the unordered image data, camera poses for all corresponding images and camera intrinsic parameters are recovered, which are both obtained by SfM \cite{schoenberger2016sfm}. 

\subsection{Fruit Segmentation}
\label{ssec:fruit_segmentation}
For fruit segmentation, two different methods are considered. 
The first is a fruit-agnostic foundation model, which offers a generalized solution applicable to all types of fruits. 
We compare this approach to a supervised neural network that has been fine-tuned specifically for apple segmentation. 
\subsubsection{Unified Fruit Model}
For the unified fruit segmentation model we used Grounded-SAM \cite{ren2024grounded}. It combines the open-set object detector Grounding DINO \cite{liu2023grounding} and the open-world segmentation model SAM \cite{kirillov2023segany}. Grounding DINO generates precise bounding boxes for every image by leveraging textual information as an input condition. The computed bounding boxes are then used by SAM as a box prompt to predict accurate segmentation masks. The advantage of this approach is that it works without fine-tuning or labeling new data.
\subsubsection{Dedicated Fruit Model}
In direct comparison, a U-Net \cite{unet} was trained on supervised data for apple segmentation. Utilizing our new data combined with the Fuji-SfM dataset \cite{fujiapple}, an apple dataset was crafted and then used for training. More information on the dataset can be found in Sec. \ref{ssec:realworlddata} and training details are listed in Sec. \ref{ssub:ppdata}.

\subsection{FruitNeRF}
\label{sec:fruit_nerf}

\emph{FruitNeRF} is the core part of the pipeline. Classic 3D representations such as point clouds, voxels, and SDFs are limited in their ability to represent fine details in complex typologies efficiently. Thus, NeRF, with its implicit volumetric representation, can store multi-view consistent multi-modal scenes such as appearance and semantic data. The network structure of \emph{FruitNeRF} is depicted in Fig. \ref{fig:FruitNeRFarchitecture}.

\subsubsection{Volumetric Rendering}

NeRF \cite{nerf} optimizes a neural radiance field by using a set of posed images and a camera model.
The scene itself is implicitly represented by a multi-layer perceptron (MLP). 
The network learns how to map a spatial coordinate point $\mathbf{x} \in \mathbb{R}^3$ and a view direction $\mathbf{d} \in \mathbb{S}^2$ to a volume density $\sigma$ and an RGB radiance $\mathbf{c} = (r, g, b)$. 
The density Field $\mathcal{F}_\sigma:\mathbf{x} \rightarrow \sigma$ is a function of the 3D position and the appearance field $\mathcal{F}_{\mathbf{c}}:(\mathbf{x}, \mathbf{d}) \rightarrow \mathbf{c}$ is a function of the 3D position $\mathbf{x}$ and view direction $\mathbf{d}$. 

The color $\mathbf{c}$ of a pixel is computed by querying the MLP at sample points along a camera ray $\mathbf{r}(t) =\mathbf{o} + t\mathbf{d}$. The ray originates in the camera $\mathbf{o}$ and its direction is determined by the pixel position.
The estimated color $\hat{\mathbf{C}}(\mathbf{r})$ for one pixel is then computed by accumulating density and color values along $K$ sampled points over the ray $\mathbf{r}$ using volumetric rendering:
\begin{equation}
\begin{split}
 & \hat{\mathbf{C}}(\mathbf{r}) = \sum_{k=1}^K \hat{T}(t_k) \alpha(\sigma(t_k)\delta_k)\mathbf{c}(t_k), \\
 \text{ where } \hspace{0.5em} & \hat{T}(t_k) = \exp\left(-\sum_{a=1}^{k-1}\sigma(t_a)\delta_a\right).
 \end{split}
\end{equation}
$\delta_k = t_{k+1} - t_{k}$ is defined as distance between two adjacent sampled points and $\alpha(x) = 1 - \text{exp}(-x)$ as the transmittance probability. An RGB rendering of a fruit tree can be seen on the left side of Fig. \ref{subfig:a}.
\subsubsection{Semantic Rendering}
\begin{figure}[b!]
     \centering
     \begin{subfigure}[b]{0.492\linewidth}
         \centering
         \includegraphics[width=\textwidth]{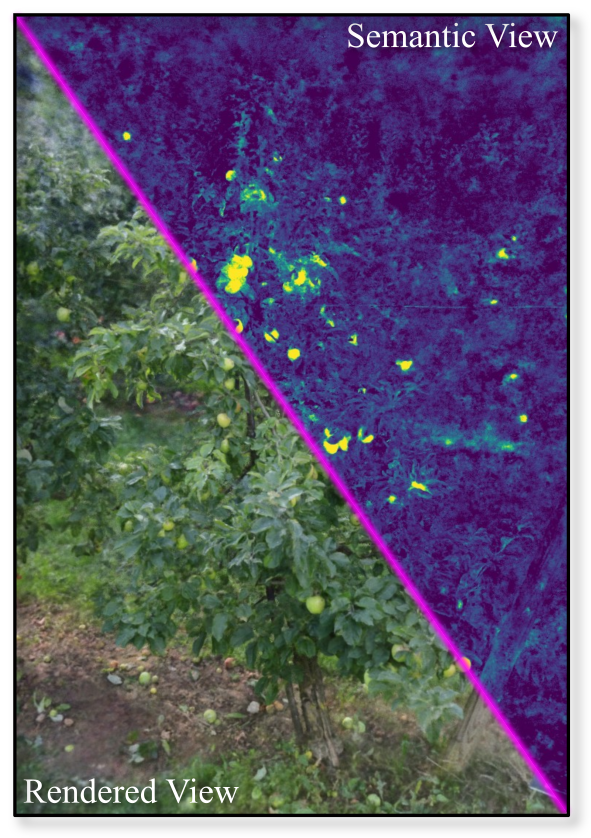}
         \caption{\emph{FruitNeRF} rendering}
         \label{subfig:a}
     \end{subfigure}
     \hfill
     \begin{subfigure}[b]{0.492\linewidth}
         \centering
         \includegraphics[width=\textwidth]{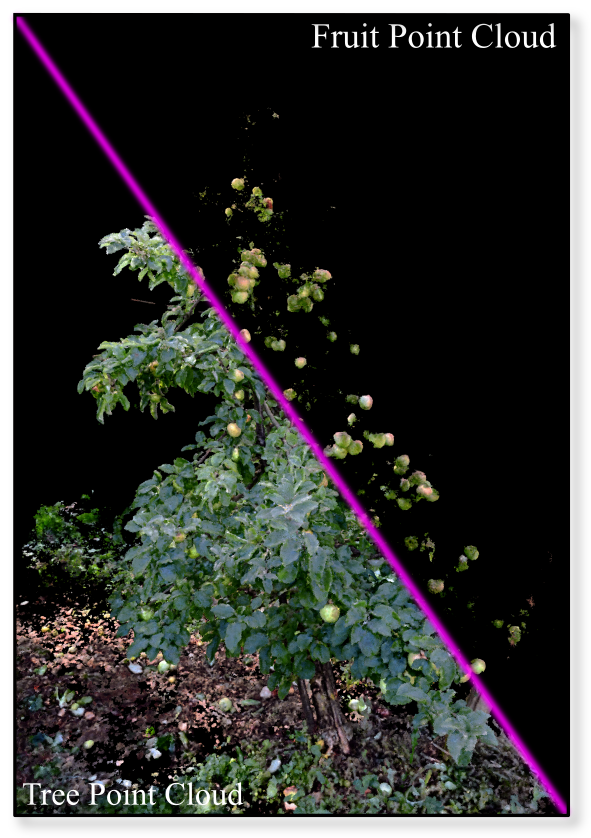}
         \caption{Extracted point clouds}
         \label{subfig:b}
     \end{subfigure}
\caption[]{Visualization of data points along the pipeline. In (a) RGB and semantic rendering are depicted. (b) shows the extracted tree point cloud from the density and the Appearance field (left) and the fruit point cloud, a combination of the Density and the Fruit Field (right).}
\label{fig:FruitNeRFRGBSem}
\end{figure}

The idea of \emph{FruitNeRF} is to encode semantic information about the fruit in 3D. 
We extend NeRF, similar to the work of Zhi \emph{et al.} \cite{zhi2021inplace}, by learning to map a 3D point not only to density and color but also to extend it to semantics. 
Therefore, an additional MLP is defined, which can be seen as a semantic field $\mathcal{F}_s:\mathbf{x} \rightarrow s$ that approximates the semantic logits as a function of only the 3D position $\mathbf{x}$. 
To approximate the estimated semantic logit $\hat{\mathbf{S}}(\mathbf{r})$ for a pixel, we accumulate the density and semantics along $K$ points sampled over the ray $\mathbf{r}$ using numerical quadrature:
\begin{equation}
\begin{split}
 & \hat{\mathbf{S}}(\mathbf{r}) = \sum_{k=1}^K \hat{T}(t_k) \alpha(\sigma(t_k)\delta_k)\mathbf{s}(t_k).\\
 \end{split}
\end{equation}
A semantic rendering of a fruit tree can be seen on the left side of Fig. \ref{subfig:a}.

\subsection{Point Cloud Export}
\label{sec:nerf_sampling}

\emph{FruitNeRF} incorporates the spatial information of fruits within its density field. 
To effectively utilize this, the \emph{FruitNeRF} volumes are sampled to process the resulting point cloud further. 
The \emph{FruitNeRF} model comprises the Density, Appearance, and Fruit Fields as depicted in Fig.~\ref{fig:FruitNeRFarchitecture}. 
The Density Field encodes the density of a point in space, independent of whether it pertains to the trunk, foliage, ground, or fruit. 
The Appearance Field encodes the corresponding color value, while the Fruit Field contains semantic information regarding the presence of a fruit.
Sampling solely from the Fruit Field would result in a scattered point cloud as during training the semantic information updates along ray and smears the information also in empty space.
To address this, we link the semantic points with the density and allow only points with a certain density to be included in the resulting point cloud. 
Fig.~\ref{subfig:b} shows the extracted point cloud from the Density and Appearance Field (left) and the fruit point cloud (right).
Afterward, the point cloud is manually cropped to include only the tree of interest, as we aim to obtain a per-tree fruit count evaluation.

\subsection{Fruit Counting}
\label{sec:fruit_count}
To enumerate individual fruits within the extracted fruit point cloud, we have developed a cascaded two-stage clustering methodology. The initial stage undertakes coarse clustering, while the subsequent stage refines this clustering process to detect invalid and multiple fruit.

Before clustering, we pre-process the point cloud by removing noise. 
This involves filtering points within a specified radius if they lack a minimum number of neighboring points. 

In the first stage of clustering, we utilize density-based spatial clustering (DBSCAN) \cite{dbscan}. 
This method offers the advantage of not requiring prior knowledge of the number of clusters present. 
Instead, it identifies clusters based on the density of closely packed points, defining clusters as regions with a high concentration of data points. 
Consequently, we identify three types of clusters: single, multi, and tiny. 
The clusters are visualized in Fig. \ref{fig:FruitNeRF_Pipeline}. 
Single-fruit clusters are directly assigned to the count through their clear identification by a similar volume to the template fruit.
Multi-fruit clusters contain more than one fruit in a packed vicinity and are identified through oversized volume.
Tiny fruit clusters are small in volume and may represent noise from erroneous segmentation or fruits captured from only a sparse set of viewpoints.

Before proceeding to the second clustering stage, we examine the set of tiny fruit clusters. 
If the distance between neighboring cluster centers is smaller than the average radius of a fruit, these clusters likely represent the same fruit, and thus, are merged. 
For the remaining tiny fruit clusters, we assess if their volume is similar to the expected volume of a target fruit. If not, we discard the cluster.

\begin{figure}[t!]
    \centering
    \includegraphics[width=\linewidth]{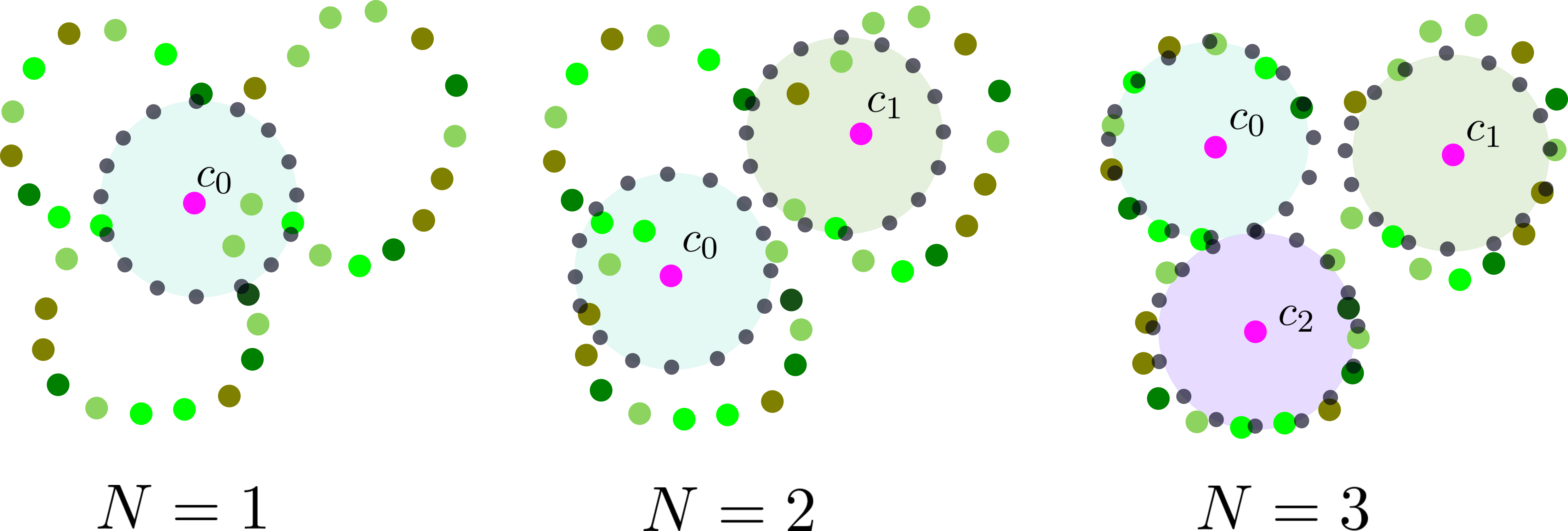}
    \label{subfig:clustering_pcd}
  \vspace{-0.2cm}
\caption[]{Second clustering stage: For multi-fruit cluster with three fruits, we simultaneously compute several cluster sizes of the fruit point cloud (greenish dots). 
Each computed cluster center $c$ (magenta dots) serves as the center point for our template fruit (smaller black dots). 
The minimum Hausdorff distance between the template point cloud and the fruit point cloud determines the number of clusters.}
\label{fig:clustering_pcd}
\end{figure}

The second clustering stage aims to determine the quantity of the multi-fruit clusters. 
E.g., specific for apples, a reasonable upper bound for the number of  apples within a cluster is $N=6$ \cite{FroutCountingCNN}. 
A multi-fruit cluster is identified if the volume of a cluster exceeds the size of our template fruit. 
In the second stage, we employ agglomeration clustering \cite{scikitlearn}, a hierarchical clustering method. 
The point cloud undergoes clustering multiple times, with a predefined cluster size ranging from 1 to $N$. 
For each clustering result, we compute the cluster center and overlay a template point cloud of the fruit. 
Subsequently, we compute the maximum mismatch between two point sets with the Hausdorff distance \cite{hausdorff_distance} through
\begin{equation}
	d_{HD}(\mathcal X,\mathcal Y) = \sup\left\{\sup_{x\in\mathcal X}\inf_{y\in\mathcal Y} d(x,y), \sup_{y\in\mathcal Y}\inf_{x\in\mathcal X} d(x,y) \right\}.
\end{equation}
It computes the distance between the point cloud of the template fruit, $\mathcal X$, and the cluster point cloud hull, $\mathcal Y$. 
We then determine the minimal distance $d_{min} = \text{min} (d_{HD}^1, \dots, d_{HD}^{N-1}, d_{HD}^N)$ and choose the corresponding cluster size. 
The steps are visualized in Fig. \ref{fig:clustering_pcd}.
\section{Experiments and Results}
\subsection{Dataset}
\label{sec:dataset}

For our experiments, we generated a series of synthetic scenes featuring fruit trees, complemented by recordings of three real-world apple trees from an orchard setting. The data has been made publicly available, and visualizations can be accessed on the project website.
\begin{figure}[b!]
    \centering
    \subfloat[][Synthetic dataset - apple, pear, plum, mango, lemon, peach]
    {\includegraphics[width=\linewidth]{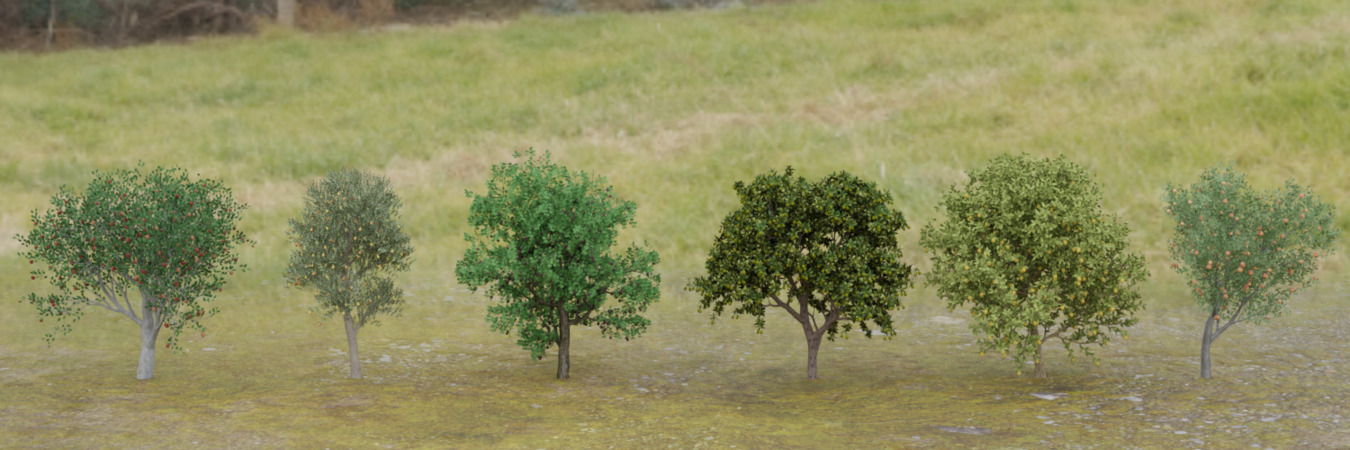}
    \label{subfig:syntheticdataset}}\vspace{0.15cm}

  \subfloat[][Real-world dataset (rendered) - Tree 03, 02, 01]{\includegraphics[width=\linewidth]{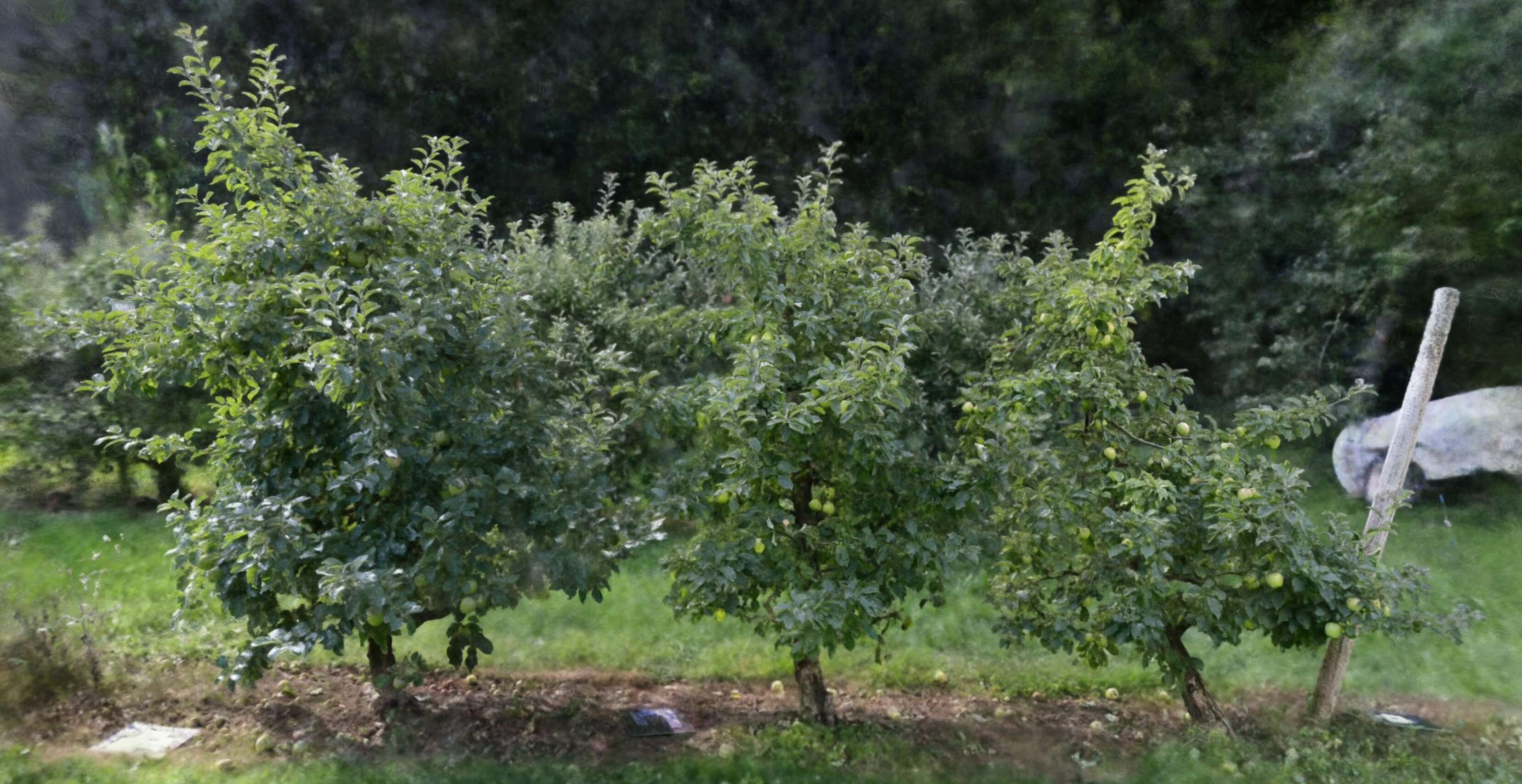} 
  \label{subfig:realdataset}}

\caption[]{Visualization of the synthetic data rendered with Blender (a) and the three apple trees of our real-world dataset (b). }
\label{fig:dataset}
\end{figure}
\subsubsection{Synthetic Blender Dataset}

The synthetic dataset was generated using Blender, with various fruit tree models, such as apple, plum, lemon, pear, peach, and mango trees, sourced from XFrog \cite{fruitassets}. 
Each tree was rendered individually using the BlenderNeRF plugin \cite{BlenderNeRF}. 
This plugin facilitated the placement of a virtual camera on a hemisphere, allowing us to extract both the camera's extrinsic and intrinsic parameters for randomly sampled perspectives directed toward the tree.

For the virtual camera, we opted for a focal length of $35~\text{mm}$ and set the image size to $1024~\text{px} \times 1024~\text{px}$. 
In addition to rendering the photometric images, we also generated semantic masks for each fruit tree. 
Furthermore, we extracted masks from Grounded-SAM for every image.
For every fruit tree, a total number of 300 images were rendered. A visualization of all fruit trees is depicted in Fig. \ref{subfig:syntheticdataset}.

\subsubsection{Real World Dataset}
\label{ssec:realworlddata}
Recordings for the real-world dataset were conducted at the Hiltpoltstein Fruit Information Center in Bavaria, Germany. 
We selected three apple trees of the \textit{Resista} variety, and their pruning closely resembles traditional fruit-growing methods. A visualization of the apple trees is shown in Fig. \ref{subfig:realdataset}.

The dataset was captured using a Nikon D7100 DSLR camera with a lens featuring a focal length of $35~\text{mm}$. 
The captured images have a resolution of $4000~\text{px} \times 6000~\text{px}$. 
We captured approximately 350 images per tree from a consistent distance of 3 meters, covering multiple heights on both sides of the tree. 
For recovering the poses of the images, we utilized COLMAP \cite{schoenberger2016sfm}. 
All apples on each tree were manually counted to obtain a per-tree ground truth. 
The apple count data is summarized in Figure \ref{fig:barapple}. For the real-world dataset, we also provide the predicted masks from both Grounded-SAM and U-Net.

\subsection{Implementation Details}



\subsubsection{Fruit Segmentation}
\label{ssub:ppdata}

To generate fruit-specific segmentation masks, we employed Grounded-SAM \cite{ren2024grounded} with pre-trained weights. 
In our pursuit of optimal mask output, we experimented with various text prompts to enhance segmentation mask quality and effectively identify different fruits within each dataset. 
Overall, we achieved satisfactory results across various fruit types using the generic text prompt "\textit{fruits}".
However, for specific fruits such as apple, plum, lemon, pear, and peach, employing the fruit's name as the text prompt yielded the most accurate results. For mangos, we did not obtain good results in both cases, but instead for using the prompt 'apple'. It should be noted that using the singular of the fruit name achieved significantly better results than the plural.
Furthermore, to increase the number of segmentation results, we set the detection threshold in grounding DINO and SAM to a low-threshold value. Additionally, for the real-world dataset we segmented the images on the maximum image size. For the computation of \emph{FruitNeRF} we down-sampled the images and semantic masks to a resolution of $1000~\text{px} \times 1500~\text{px}$.

Our second network is a self-trained U-Net, using a manually annotated subset of 62 images from our real-world data. 
These images were tiled into smaller $2000~\text{px}\times2000~\text{px}$ sub-images and resized by a factor of 0.5 to meet GPU memory constraints. 
To enhance the dataset, we incorporated 2D images and segmentation masks from \cite{fujiapple} and employed a random-seeded augmentation pipeline that includes geometric transformations, color distortions, and pixel dropout. 
A PyTorch implementation of U-Net \cite{unettorch} was trained for the task of binary segmentation of apples.

One key difference between masks generated with SAM and U-Net is that SAM produces masks with soft edges that do not precisely align with the image edge e.g., between apples and leaves. 
Conversely, U-Net produces masks that closely resemble the original image, offering a more accurate representation.
Both architectures were evaluated using the validation split from the fuji dataset \cite{fujiapple} as a holdout test set. 
Segmentation results were measured using intersection over union weighted on class prevalence per image. 
Segmentation results averaged over all test set images were 0.919 for SAM and 0.962 for U-Net.
The results of \emph{FruitNeRF} based on Grounding DINO and SAM masks compared to the U-Net results for apple counting are displayed in Fig. \ref{fig:barapple}.

\begin{figure}[b!]
    \centering
    \includegraphics[width=\linewidth]{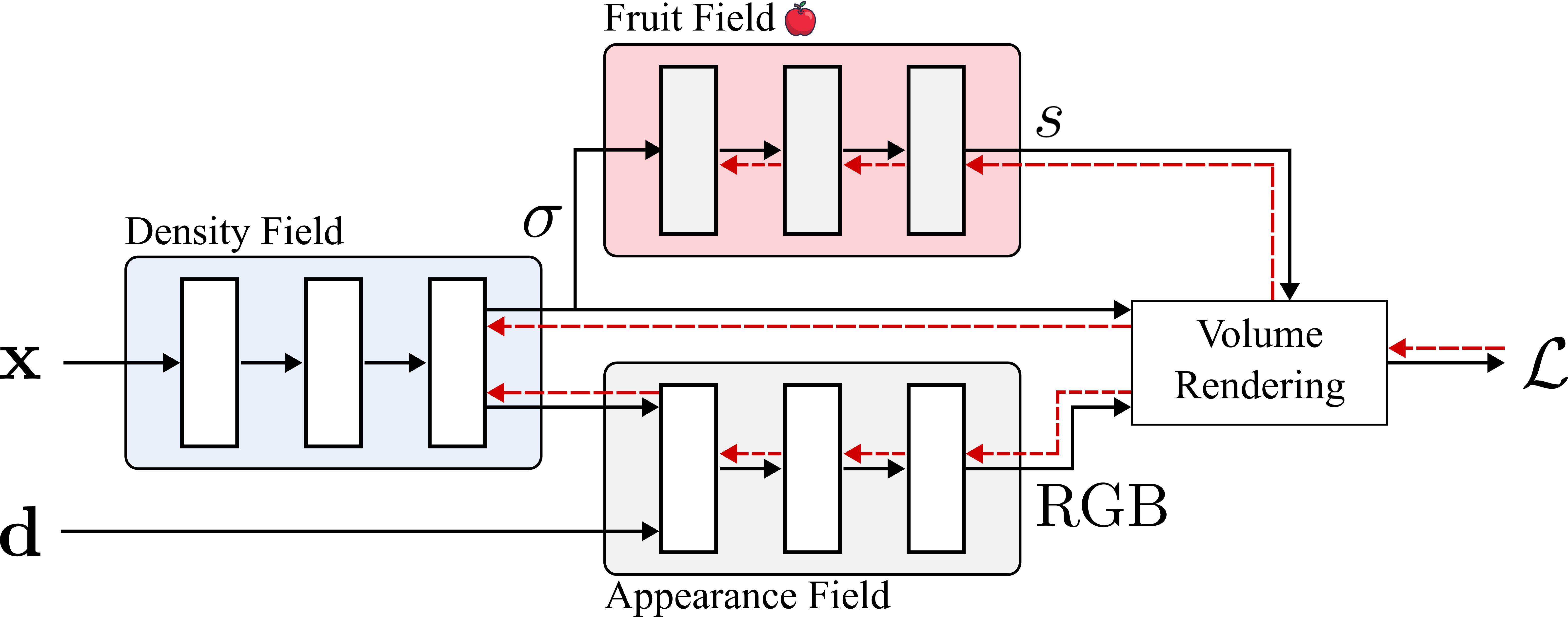}
\caption[]{Overview of the \emph{FruitNeRF} architecture, which is split up into three different components. The density field encodes the volume density $\sigma$, the Appearance Field the color $\textit{RGB}$, and the Fruit Field the semantic information about the fruit in space. The dashed red arrow indicates the flow direction of the gradient. The figure is inspired by semantic NeRF \cite{zhi2021inplace} and adapted from Özer \emph{et al.} \cite{thermalnerf}.}
\label{fig:FruitNeRFarchitecture}
\end{figure}

\begin{table*}[t!] 
\definecolor{cellgreen}{RGB}{247,203,153}
\center
\caption{Detected fruit on different fruit types. The used synthetic data have an image size of $1024~$px $\times 1024~$px and contain 300 frames sampled from the upper hemisphere. The data is listed with rendered GT semantic masks and with masks generated by SAM.}
\begin{tabular}{l || c  c  c  c  |c  c  c  c  l } 
\toprule
Fruit Type  & {GT Mask}                   & Precision    & Recall    & F1-Score   & {SAM}                     & Precision    & Recall    &   F1-Score   & Text prompt       \\ \midrule  
Apple       & \cellcolor{cellgreen} 283/283   & 1.0  & 1 & 1  & 282/283                       &  0.992  & 0.989  &  0.991  & apple         \\             
Plum        & \cellcolor{cellgreen} 651/781   & 0.973 & 0.812 & 0.885 & 315/781                       & 1.0  & 0.403 &   0.575 & apple \& plum        \\            
Lemon       &  316/326                        & 0.993 & 0.963 & 0.978 &\cellcolor{cellgreen} 326/326  & 0.982 & 0.982 &   0.982 & lemon       \\            
Pear        & \cellcolor{cellgreen} 236/250   & 1.0  & 0.944 & 0.971 & 229/250                       & 1.0  & 0.916 &   0.956 & pear         \\           
Peach       & \cellcolor{cellgreen} 148/152   & 1.0  & 0.973 & 0.987 & \cellcolor{cellgreen} 148/152 & 1.0  & 0.973 &   0.987 & peach              \\       
Mango       &  \cellcolor{cellgreen}926/1150   & 0.978 & 0.788 & 0.873 & 807/1150                       & 0.989  & 0.694 &  0.816 & apple        \\\bottomrule        
\end{tabular}
\label{tab:fruit_nerf_count}
\end{table*}  
\subsubsection{NeRF Implementation and Training}
\label{ssub:training}
The basis for \emph{FruitNeRF} is Nerfacto \cite{nerfstudio}, a method that combines state-of-the-art components from recent papers with significant impact regarding ray generation and sampling, scene contraction, and NeRF fields. 

For \emph{FruitNeRF} we extended Nerfacto by a semantic component, also referred to as Fruit Field. 
The overview of the NeRF architecture is depicted in Fig. \ref{fig:FruitNeRFarchitecture}. 
The semantic branch takes only the feature vector of the predicted density as an input and not the viewing direction, as the semantic modality is view-independent.
For training on the posed images, we used a default rendering loss to compute the photo-metric error between the pixel's RGB value $\mathbf{C}(\mathbf{r})$ and the predicted color value $\mathbf{\hat{C}(\mathbf{r})}$ for ray $\mathbf{r} $ by:
\begin{equation}
	\mathcal{L}_{\text{Photo}} = \frac{1}{|\mathcal{R}|}\sum_{\mathbf{r} \in \mathcal{R}}||\mathbf{C}(\mathbf{r}) - \mathbf{\hat{C}}(\mathbf{r})||_2^2,
\end{equation}
where $\mathcal{R}$ is denoted as the set of sampled rays. Regarding the semantics, we leveraged binary cross entropy for pixel-wise classification probability in fruit or background class by:
\begin{equation}
	\mathcal{L}_{\text{Sem}} = \frac{1}{|\mathcal{R}|}\sum_{\mathbf{r} \in \mathcal{R}}p(\mathbf{r})\log \hat{p}(\mathbf{r}) + (1- p(\mathbf{r}))\log (1-\hat{p}(\mathbf{r})).
\end{equation}
The total training loss is then composed by:
\begin{equation}
	\mathcal{L} = \mathcal{L}_{\text{Photo}} + \mathcal{L}_{\text{Sem}}.
\end{equation}
In semantic NeRF \cite{zhi2021inplace} the semantic loss gets an additional weight as they propagate the semantic gradient back through the Density Field. 
We on the other side restrict back-propagation of the semantic loss to only the Fruit Field. 
Otherwise, the Density Field would focus on predicting density values for spatial points only belonging to fruit. 

We implemented two different network sizes: \emph{FruitNeRF} and \emph{FruitNeRF-Big}.
\emph{FruitNeRF} utilizes 2 layers for its Fruit Field, with a hidden layer size of 64 neurons.
The input layer has a dimension of 15 (the input is a 15-dimensional latent vector from the Density Field).
The output dimension of the neural network is 64, which serves as input for a neural segmentation head that reduces the dimension to a single class.
For \emph{FruitNeRF-Big}, we increased the layer depth to 3 and increased the hidden dimension to 128 neurons. The input dimension is set to 30 and the output size is kept identical. In the larger variant, we also increased the overall capacity for the density and appearance according to the Nerfacto-Big implementation \cite{nerfstudio}. 
For training, we used an Nvidia RTX A5000 with 24GB VRAM. Training time for \emph{FruitNeRF} is an estimate of $12$ min and for \emph{FruitNeRF-Big} roughly $2$h and $30$min. We used an input image size of $1000~\text{px} \times 1500~\text{px}$ for real-world data and $1024~\text{px} \times 1024~\text{px}$ for synthetic data.



\subsubsection{Field Export}
\label{ssub:export}

A sampling of the \emph{FruitNeRF} volume is achieved by viewing the scene with an orthographic camera model.
We first define a unit cube or use a predefined region of interest around our scene and select one side of the cube to be the image plane. By splitting the image plane into pixels, we can cast multiple rays and query the \emph{FruitNeRF} at a predefined number of steps. By discarding points with a density and semantic value under a fixed threshold, we obtain a point cloud with only fruit points.

%
%

\subsection{FruitNeRF Evaluation and Results}

In this section, we evaluate our approach using synthetic and real-world data with a focus on the following points:

\begin{itemize}
    \item  The \emph{FruitNeRF} counting performance is evaluated across six distinct types of fruit based on the synthetic dataset.
    \item  The amount and resolution of input images are varied to investigate the influence of these parameters on resulting fruit counts.
    \item  Two scaled \emph{FruitNeRF} architectures are applied to multiple apple datasets and various segmentation pipelines to demonstrate the robustness and practical applicability of our approach in a real-world setting.
\end{itemize}


For the first experiment, we trained the default \emph{FruitNeRF} model on each type of fruit using 300 images at a resolution of $1024~\text{px} \times 1024~\text{px}$. 
Afterward, we evaluated the performance using both ground truth masks and masks generated by Grounded-SAM \cite{ren2024grounded}. The summarized results are presented in Table \ref{tab:fruit_nerf_count}. 
In general, \emph{FruitNeRF} with ground truth masks exhibited an average F1-score of $0.95$ compared to SAM-generated masks at $0.88$. 
Particularly for apples, lemons, pears, and peaches, both sets of results closely matched the actual fruit count, resulting in excellent precision and recall values. The worst results with SAM were achieved for plums and mangoes with recall values of $0.4$ and $0.69$.
This performance drop can be attributed to the significantly higher fruit occlusions, which impaired SAM's prediction quality. SAM masks are more accurate for trees with relatively low fruit counts and fewer occlusions. It can be observed that the centers of the trees were either poorly reconstructed or not present in the fruit cloud at all. 
\begin{figure}[b!]
    \begin{tikzpicture}
        \begin{axis}[
            width=\linewidth,
            height=.7\linewidth,
            xlabel={Num. Frames },
            ylabel={Fruit Num.},
            xmin=0, xmax=105,
            ymin=0, ymax=510,
            xtick={5, 20, 30, 40, 50, 60, 70, 100, 150, 200},
            ytick={0,100,200,283, 400, 500},
            legend style={at={(0.85,0.47)},
                anchor=north,},
            ymajorgrids=true,
            xmajorgrids=true,
            grid style=dashed,
        ]
        \draw[line width=0.01 mm] (0,283) -- (120,283);

        \addplot[ color=magenta, mark=asterisk, mark options={solid, scale=1.3, fill=white}, dashdotdotted, line width=0.25mm]coordinates {(5,0)(10,3)(15,0)(20,203)(30,292)(40,286)(50,282)(60,284)(70,287)(100,284)};
        
        \addplot[ color=blue, mark=square*, mark options={solid, fill=white}, dashed, line width=0.25mm]coordinates {(5,0)(10,6)(15,0)(20,225)(30,301)(40,288)(50,288)(60,287)(70,283)(100,284)};

        \addplot[ color=red, mark=diamond*, mark options={solid, scale=1.3, fill=white}, line width=0.25mm]coordinates {(5,0)(10,2)(15,0)(20,392)(30,332)(40,292)(50,288)(60,286)(70,283)((100,283)};
        
        \addplot[ color=black, mark=triangle*, mark options={solid, scale=1.3, fill=white}, dotted, line width=0.25mm]coordinates {(5,0)(10,2)(15,0)(20,696)(30,497)(40,356)(50,313)(60,291)(70,293)(100,283)};

         \draw[-latex] (axis cs:45,450) -- (axis cs:45,490);
         \addplot[color=black, mark=triangle*, mark options={solid, scale=1.3, fill=white}, dotted, line width=0.25mm]
         coordinates{
         (48.5, 466)
         };
         \node[right] at (axis cs:49.5,466) {\footnotesize $(20, 696)$};


        \legend{$1024$, $512$, $256$, $128$}
        \end{axis}
    \end{tikzpicture}
    \caption{Fruit count dependence on number of frames. For this evaluation, we utilized the ground truth mask with different sizes. The ground truth count of the synthetic apple tree is $283$.}

    \label{fig:fruit_nerf_appletree_eval}
\end{figure}
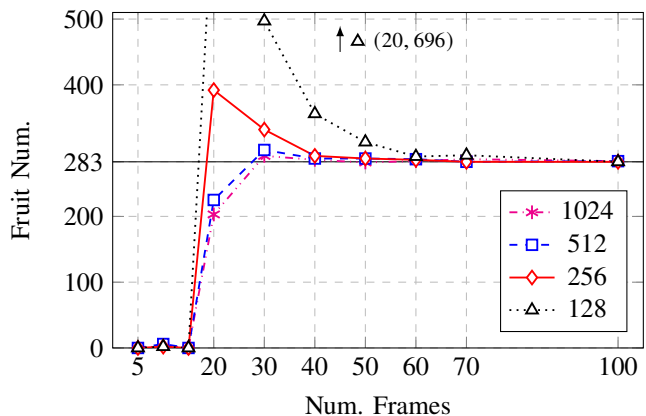

\pgfplotstableread[row sep=\\,col sep=&]{
    interval   & SAM  & SAM-B & U-Net & U-Net-B & GT   \\
    Tree $01$  & 147  & 173   & 146   & 172     &  179 \\
    Tree $02$  & 86  & 112   & 88   & 114     & 113 \\
    Tree $03$  & 190  & 264   & 255   & 243     &  291 \\
    }\mydata

\pgfplotstableread[row sep=\\,col sep=&]{
    interval   & SAM  & SAM2 & SAM-B & SAM-B2 & U-Net & U-Net2  & U-Net-B & U-Net-B2 & GT     & GT2  \\
    Fuji       & 842  & 974  & 1191  & 1556   & 1059  & 1323    &  1142  &  1459    &   1455 &   0  \\
    }\myfujidata

\begin{figure*}[t!]
        \begin{tikzpicture}[baseline=0pt]
        \definecolor{clr1}{RGB}{204, 221, 113}
        \definecolor{clr2}{RGB}{50, 254, 200}
        \definecolor{clr3}{RGB}{90, 178, 214}
            \begin{axis}[
                    ybar,
                    bar width=0.5cm,
                    enlarge x limits=0.3,
                    width=0.72\linewidth,
                    height=.35\linewidth,
                    legend style={nodes={scale=0.7, transform shape},
                        at={(0.372, 1)},
                        anchor=north,legend columns=-1, style={ column sep=.15cm}},
                    symbolic x coords={Tree $01$, Tree $02$, Tree $03$, Fuji},
                    xtick pos=bottom,
                    ytick pos=left,
                    xtick=data,
                    nodes near coords,
                    nodes near coords align={vertical},
                    every node near coord/.append style={font=\tiny},
                    ymajorgrids=true,
                    ymin=0,ymax=325,
                    grid style=dashed,
                    ytick={0,50, 100, 150, 200, 250, 300},
                    ylabel={Fruit Num.},
                ]
                \addplot [black, fill=clr1, postaction={pattern=north east lines}]  table[x=interval,y=SAM] {\mydata};
                \addplot [black, fill=clr1, postaction={pattern=north west lines}]  table[x=interval,y=SAM-B] {\mydata};
                \addplot [black, fill=clr2, postaction={pattern=dots}] table[x=interval,y=U-Net]{\mydata};
                \addplot [black, fill=clr2, postaction={pattern=grid}] table[x=interval,y=U-Net-B]{\mydata};
                \addplot [black, fill=clr3, postaction={pattern=horizontal lines}] table[x=interval,y=GT]{\mydata};
                \legend{SAM, SAM-B, U-Net, U-Net-B, GT}
            \end{axis}
        \end{tikzpicture}
        \begin{tikzpicture}[
        baseline=0pt,
         every axis/.style={
                    ybar,
                    enlarge x limits=0.3,
                    width=0.3\linewidth,
                    height=.35\linewidth,
                    symbolic x coords={Fuji},
                    xtick pos=bottom,
                    ytick pos=right,
                    xtick=data,
                    nodes near coords,
                    nodes near coords align={vertical},
                    every node near coord/.append style={font=\tiny},
                    ymin=0, ymax=1700,
                    ymajorgrids=true,
                    grid style=dashed,
                    ytick={0, 500, 1000, 1455},
                    bar width=0.5cm,
         },
        ]
        \definecolor{clr1}{RGB}{204, 221, 113}
        \definecolor{clr2}{RGB}{50, 254, 200}
        \definecolor{clr3}{RGB}{90, 178, 214}
            \begin{axis}[
                ]                
                \addplot [black, fill=white]  table[x=interval,y=SAM2] {\myfujidata};
                \addplot [black, fill=white]  table[x=interval,y=SAM-B2] {\myfujidata};
                \addplot [black, fill=white] table[x=interval,y=U-Net2]{\myfujidata};
                \addplot [black, fill=white] table[x=interval,y=U-Net-B2]{\myfujidata};
                \addplot [black, fill=white] table[x=interval,y=GT2]{\myfujidata};
            \end{axis}
            \begin{axis}[
                    nodes near coords,
                    nodes near coords align={vertical},
                    every node near coord/.append style={font=\tiny},
                ]                
                \addplot [black, fill=clr1, postaction={pattern=north east lines}]  table[x=interval,y=SAM] {\myfujidata};
                \addplot [black, fill=clr1, postaction={pattern=north west lines}]  table[x=interval,y=SAM-B] {\myfujidata};
                \addplot [black, fill=clr2, postaction={pattern=dots}] table[x=interval,y=U-Net]{\myfujidata};
                \addplot [black, fill=clr2, postaction={pattern=grid}] table[x=interval,y=U-Net-B]{\myfujidata};
                \addplot [black, fill=clr3, postaction={pattern=horizontal lines}] table[x=interval,y=GT]{\myfujidata};
                \end{axis}
        \end{tikzpicture}

        \caption{The estimated apple count for the 3 recorded real-world datasets was evaluated using masks generated with SAM and U-Net, employing two different \emph{FruitNeRF} sizes: default and big (-B). The image size for each tree dataset is $1000$~px $\times$ $1500$~px. GT apple counts (Tree 01-03) were obtained through manual counting on-site, and clustering parameters were kept consistent per tree. The Fuji dataset \cite{fujiapple} contains 11 trees. The highlighted bar indicates the correctly counted (recall), and the white bar on top is the number of overall counted fruit.}
        \label{fig:barapple}
\end{figure*}
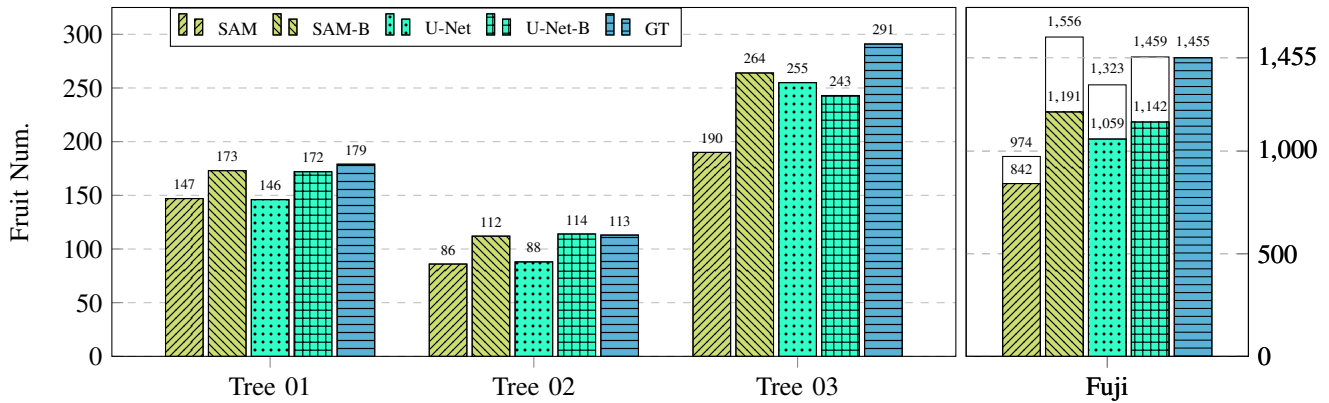


In our second experiment, we evaluated the impact of varying numbers of images and image resolutions. 
To achieve this, we down-sampled both RGB images and semantic masks from the original apple tree rendering to resolutions of  $1024$, $512$, $256$, and $128$~px.
We initiated the experiment with a consistent set of $5$ images with ground truth masks and incrementally introduced additional images up until $100$. 
For each set of images, we trained a new \emph{FruitNeRF}, extracted the fruit point cloud, and applied our fruit count clustering with constant parameters.
The evaluation results for different resolutions are presented in Figure \ref{fig:fruit_nerf_appletree_eval}.

It is evident that \emph{FruitNeRF} struggles to learn a meaningful representation with a sparse number of images, particularly when fewer than $20$ images are used. 
A notable improvement is observed when using 20-30 images with fruit count peaks across all resolutions. 
This phenomenon is caused by insufficient information presented to the Fruit Field, leading to the smearing of semantic knowledge over the entire tree, resulting in a significant increase in clusters falsely counted as fruits. 
As \emph{FruitNeRF} is exposed to more images, the precision of the fruit cloud improves, as well as the accuracy of the counting. 
To achieve accurate fruit counts, we found that for a resolution of $128$~px, 60 images are required; for $256$~px, 50 images are needed; and for $512$~px and $1024$~px, 30-40 images are necessary.


Lastly, we evaluated \emph{FruitNeRF} using two different real-world datasets. We used masks generated by Grounded-SAM and our U-Net.
Additionally, we employed a normal (\emph{FruitNeRF}) and a larger NeRF model (\emph{FruitNeRF}-Big). 

For our dataset, depicted in Fig. \ref{fig:barapple} on the left, we achieved a detection rate on average of $\sim89\%$ for Trees 1 and 2 for both U-Net and SAM.
However, results for Tree 3 reveal a detection rate drop to $\sim82\%$, which is caused by the tree's more complex structure and resulting extensive occlusions within the tree crown. 

As a second dataset, we choose the Fuji-SfM dataset, which consists of 11 trees in a row captured with $582$ images from both tree sides.
\emph{FruitNeRF-Big} with SAM and U-Net generated masks achieves a F1-Score of $0.79$ and $0.78$ respectively, which is near the $0.88$ of the original paper \cite{fuji_count}. 
Those are decent results, considering that our pipeline offers unified counting of arbitrary fruits and was not fine-tuned on apples specifically. 
For the Fuji data, the smaller models performed worse than those with more parameters. 
The gap between the differently sized architectures can be attributed to the elongated shape of the Fuji scene, which does not make use of the space efficiently. 
Larger models have increased capacity that can be effectively used to encode data of complex scenes more precisely. 
In comparison, our recorded apple trees (Tree 01-03) can make better use of the volume within the unit cube as their scenes are less complex and thus, result in smaller performance gaps between \emph{FruitNeRF-Big} and \emph{FruitNeRF}.

\section{Limitations}

While \emph{FruitNeRF} demonstrates promising results and addresses common challenges in fruit counting, several limitations persist. 
The primary constraint lies in the significant training time and GPU memory requirements, as our method heavily relies on image data coverage and quality. 
Consequently, our approach is not yet suitable for real-time applications or edge computing. 
Initial experiments with Gaussian Splatting \cite{gssplatting} have decreased computing times for our pipeline, and utilizing other architectures, such as PAgNerf \cite{pagnerf}, might lead to even faster computations.

Moreover, the cascaded clustering technique involves hyperparameters that require manual adjustment according to the type of fruit, complicating unified automation processes. 
Additionally, the second clustering stage is bound to a nominal fruit size, necessitating adaptation for valid counting of varying fruit sizes at different growth stages. 
Implementing a learned approach for fruit detection within the point cloud could significantly alleviate this issue and enhance the efficiency of fruit counting.


As an industrial application, low lighting conditions and different exposure levels due to changing weather conditions could worsen the results of the NeRF reconstruction. 
Therefore, implementations such as Low-Light NeRF \cite{llnerf} and HDR-NeRF \cite{hdrnerf} could tackle these problems. 
Additionally, the presence of a non-static scene, caused by wind, must be further investigated to determine if dynamic NeRF approaches such as RobustNeRF \cite{robustnerf} can solve these issues.

Nevertheless, we are confident that the application potential of our approach will benefit from rapid technological improvements in hardware and 3D reconstruction techniques, which will mitigate these limitations.

\section{Conclusion and Future Work}

This paper introduces \emph{FruitNeRF}, a novel framework designed to accurately count visible fruits within a neural radiance field. 
Leveraging only 2D images, \emph{FruitNeRF} facilitates precise 3D reconstructions of trees, enabling accurate fruit counting. 
By integrating Grounded-SAM into the \emph{FruitNeRF} pipeline, arbitrary types of fruit can be counted without the need for costly annotations and the training of a U-Net.

Our framework has been thoroughly validated in both synthetic and real-world scenarios, representing the first application of NeRFs for fruit counting to our knowledge.
The experiments demonstrate that \emph{FruitNeRF} achieves an F1-score of $0.95$ on our synthetic dataset with ground truth masks, while masks generated by SAM attain around an F1-score of $0.88$ averaged over six different fruit species.
Additionally, we have illustrated that excellent results can be obtained with only 40 images per tree and a resolution of  $512~\text{px} \times 512~\text{px}$, demonstrating the scalability and effectiveness of our approach.
Fruit counting on our self-recorded real-world apple dataset showcases a detection rate exceeding $89\%$ across various masks and network architectures. 
For the Fuji benchmark dataset, we demonstrate an F1-score of $0.79$.

From this novel fruit-counting approach, several promising directions for future research emerge. 
Primary efforts should focus on improving the clustering step, as it is highly sensitive to hyper-parameter tuning.
Additionally, to broaden the applicability of \emph{FruitNeRF}, the use of time-series images should be considered. This could reduce the number of required images and help achieve real-time performance.

As we aim to be a unified fruit counting network, further investigation into soft fruits such as strawberries, raspberries, grapes, and other small-sized fruits should be carried out.

Extending the framework's utility beyond individual trees to entire orchard rows could be explored through the integration of online pose estimation methods like Simultaneous Localization and Mapping (SLAM). 
Simulation environments, such as SLAM in Blender \cite{blenderslam}, could serve as valuable test beds for refining detection rate.

Moreover, the versatility of neural radiance fields extends beyond visible light, presenting opportunities to incorporate other modalities such as near-infrared or thermal imaging, as demonstrated by \"Ozer \cite{thermalnerf}. 
\emph{FruitNeRF} demonstrates the potential to enhance fruit counting capabilities, but leveraging neural radiance fields also enables broader orchard analysis, including ripeness assessment, stress level monitoring, and guidance of harvesting robots.


\section{Acknowledgement}
We extend our gratitude to \textbf{Adam Kalisz} for his unique Blender skills, \textbf{Victoria Schmidt}, and \textbf{Annika Killer} for their invaluable assistance in evaluating the recorded apple trees. Additional thanks to \textbf{Roland Gruber}, \textbf{Christoph Drescher}, and \textbf{Julian Bittner} for their annotation support.

This project is funded by the 5G innovation program of the German Federal Ministry for Digital and Transport under the funding code 165GU103B.


\end{document}